\documentclass{article}

\usepackage[numbers]{natbib}
\usepackage[final]{neurips_2019}

\usepackage[utf8]{inputenc}
\usepackage[T1]{fontenc}
\usepackage{hyperref}
\usepackage{url}
\usepackage{booktabs}
\usepackage{amsfonts}
\usepackage{nicefrac}
\usepackage{microtype}
\usepackage{graphicx}
\usepackage{xcolor}
\usepackage{lipsum}

\usepackage{adjustbox}
\usepackage{array}
\usepackage{flexisym}
\usepackage[inline]{enumitem}
\usepackage{bm}
\usepackage{subcaption}

\newcommand{\rot}[1]{\multicolumn{1}{l}{\adjustbox{angle=45,lap=\width-1em}{#1}}}

\graphicspath{{figures/}}

\title{Gestalt: a Stacking Ensemble for SQuAD2.0}

\author{
  Mohamed El-Geish \\
  Department of Computer Science \\
  Stanford University \\
  \texttt{elgeish@stanford.edu}
}

\begin{document}
\maketitle

\begin{abstract}
  We propose a deep-learning system --- for the SQuAD2.0 task --- that finds, or indicates the lack of, a correct answer to a question in a context paragraph. Our goal is to learn an ensemble of heterogeneous SQuAD2.0 models that, when blended properly, outperforms the best model in the ensemble per se. We created a stacking ensemble that combines top-N predictions from two models, based on ALBERT and RoBERTa, into a multiclass classification task to pick the best answer out of their predictions. We explored various ensemble configurations, input representations, and model architectures. For evaluation, we examined test-set EM and F1 scores; our best-performing ensemble incorporated a CNN-based meta-model and scored 87.117 and 90.306, respectively --- a relative improvement of 0.55\% for EM and 0.61\% for F1 scores, compared to the baseline performance of the best model in the ensemble, an ALBERT-based model, at 86.644 for EM and 89.760 for F1.
\end{abstract}

\section{Introduction}
Despite the simplicity of the idea, ensemble learning has been widely successful in a plethora of tasks --- ranging from machine learning contests to real-world applications~\cite{zhang2012ensemble}. We aim at using ensemble learning to create a deep-learning system for a machine reading comprehension application: question answering (QA). The main goal of QA systems is to answer a question that's posed in a natural manner using a human language. In this research area, we find many examples for variants of the QA task~\cite{eisenstein2019introduction}. Closed-domain QA systems answer questions about a specific domain while open-domain ones answer questions about a myriad of topics~\cite{yang-etal-2015-wikiqa, ferrucci2010build}. Knowledge-base QA systems find answers in a specific knowledge base such as Freebase~\cite{bollacker2008freebase}. Multiple-choice QA systems pick an answer out of multiple choices like in the MCTest~\cite{richardson-etal-2013-mctest} and RACE~\cite{lai-etal-2017-race} tasks. Some QA systems directly answer a given question by generating complete sentences like in ELI5~\cite{fan-etal-2019-eli5} while others extract a short span of text in a corresponding context paragraph to present it as an answer; the latter is the main objective of SQuAD (the Stanford Question Answering Dataset) models~\cite{rajpurkar-etal-2016-squad}. In SQuAD2.0, an additional challenge was introduced: The model has to indicate when a question is unanswerable given the corresponding context paragraph --- see Figure~\ref{fig:squad2.0} for SQuAD2.0 examples.

\begin{figure}[ht]
    \centering
    \begin{tabular}{p{0.9\linewidth}}
        \toprule
        \textbf{Paragraph:} "Bethencourt took the title of King of the Canary Islands, as vassal to Henry III of Castile. In 1418, Jean's nephew Maciot de Bethencourt sold the rights to the islands to Enrique Pérez de Guzmán, 2nd Count de Niebla." \\ \\
        \textbf{Question 1:} "What title did Henry II take in the Canary Island?" \\
        \textbf{Ground Truth Answer:} <No Answer> \\
        \textbf{Plausible Prediction:} \color{red}{King of the Canary Islands} \\ \\
        
        \textbf{Question 2:} "Who bought the rights?" \\
        \textbf{Ground Truth Answer:} Enrique Pérez de Guzmán \\
        \textbf{Plausible Prediction:} \color{blue}{Pérez} \\
        \bottomrule
    \end{tabular}
    \vspace{1em}
    \caption{Two examples of SQuAD2.0 questions and answers written by crowdworkers, along with plausible model predictions: the one in red is incorrect while the one in blue is incomplete.}
    \label{fig:squad2.0}
\end{figure}

SQuAD2.0 systems face many challenges: The task requires accurately concocting some forms of natural language representation and understanding that aid in processing a question and the context to which it relates, then selecting a reasonable correct answer that humans may find satisfactory or indicate the lack of such answer. The vast majority of modern systems, which outperform humans according to the SQuAD2.0 leaderboard~\cite{leaderboard}, try to find two indices: the start and end positions of the answer span in the corresponding context paragraph (or sentinel values if no answer was found). Recently, this has been usually done with the aid of Pre-trained Contextual Embeddings (PCE) models, which help with language representation and understanding~\cite{devlin-etal-2019-bert, yang2019xlnet, sanh2019distilbert, liu2019roberta, 2019t5, Lan2020ALBERT}. The SQuAD2.0 leaderboard shows that ensembles improve upon the performance of single models: In~\cite{devlin-etal-2019-bert}, BERT introduced an ensemble of six models that added 1.4 F1 points; ALBERT's ensemble averages the scores of 6 to 17 models, leading to a gain of 1.3 F1 points, compared to the single model~\cite{githubALBERT}; RoBERTa and XLNet also introduced ensembles but did not provide sufficient details~\cite{liu2019roberta, yang2019xlnet}. Our QA ensemble system added a gain of 0.473 EM points (0.55\% relative gain) and 0.546 F1 points (0.61\% relative gain), compared to the best-performing single model in the ensemble, when measured using the project's test set (which is different from the official SQuAD2.0 hidden test set).

A key difference in our approach is the use of stacking~\cite{wolpert1992stacked} to combine top-$N$ predictions produced by each model in the ensemble. We picked heterogeneous PCE models, fine-tuned them for the SQuAD2.0 task, and combined their top-$N$ predictions in a multiclass classification task using a convolutional neural meta-model that selects the best possible prediction as the ensemble's output. Since each model in the ensemble is learned in an idiosyncratic manner, we expect their results (given the same input) to vary --- a behavior that's analogous to asking humans, who come from diverse backgrounds, for their opinions.

\section{Related Work}
Pre-trained Contextual Embeddings (PCE) models have been instrumental in advancing language representation learning and achieving state-of-the-art results in many NLP tasks; the Bidirectional Encoder Representations from Transformers (BERT) family of models have been at the forefront~\cite{devlin-etal-2019-bert}. ALBERT, A Lite BERT, has been notably excelling at the SQuAD2.0 task~\cite{Lan2020ALBERT} --- especially when used in an ensemble: At the time of writing, the top 7 models on the SQuAD2.0 leaderboard are ALBERT-based ensembles~\cite{leaderboard}. In ~\cite{Lan2020ALBERT}, the ALBERT ensemble selects model checkpoints with the best development set performance then averages their prediction scores to produce an aggregate answer for a SQuAD2.0 question (or to indicate the lack of an answer). ALBERT's ensemble achieved an F1 score of 92.2 while the single-model result was 90.9 (after 1.5M training steps).

One challenge of averaging prediction scores is giving each model in the ensemble an equal voting power regardless of its performance; another is susceptibility to outliers --- a single outlier can tip the scales easily. There are more sophisticated ways of combining models to address such challenges, including assigning each vote a weight; we may consider the aforementioned methods special cases of stacked generalization~\cite{wolpert1992stacked}. Stacking (another name for stacked generalization) works by learning a meta-model that combines the predictions of various models in order to produce predictions that minimizes the generalization error. A stacking ensemble exploits the independence of heterogeneous models, which were built differently, since the probability of them colluding to give wrong answers is minimal~\cite{witten2016data}.

Stacking ensembles for SQuAD2.0 are to benefit from blending heterogeneous PCE models like XLNet~\cite{yang2019xlnet}, DistilBERT~\cite{sanh2019distilbert}, RoBERTa~\cite{liu2019roberta}, Text-To-Text Transfer Transformer (T5)~\cite{2019t5}, and ALBERT~\cite{Lan2020ALBERT} --- each of which has its own unique contributions. While BERT learns using Masked Language Modeling and Next Sentence Prediction (NPS) loss, XLNet maximizes the expected likelihood over all permutations of words in a sentence (Permutation Language Modeling). DistilBERT is a distilled version of BERT that's cheaper and faster to fine-tune while retaining 97\% of its language understanding capabilities. RoBERTa improved upon BERT's performance thanks to careful hyperparameter choices, longer training using more data that include longer sequences, and removing the NPS objective. T5 scaled up model size (up to 11B parameters) and learned from 120B words --- the Colossal Clean Crawled Corpus (C4) dataset it introduced in~\cite{2019t5}. ALBERT improved upon BERT in terms of parameter efficiency and the use of inter-sentence coherence loss instead of NPS loss.

In the same manner ensembles showed significant improvements in performance for tasks in ~\cite{wolpert1992stacked, zhang2012ensemble} (e.g., the NETtalk task of translating text to phonemes), we propose approaches below to create a stacking ensemble for the SQuAD2.0 task with the performance improvement goal in mind.

\section{Approaches}
\label{sec:approaches}
Creating a stacking ensemble for SQuAD2.0 entails building a pipeline of two stacked stages: level 0 and level 1. In level 0, models learn from SQuAD2.0 and produce predictions, which are then used as input to level 1, for a meta-model, to produce better predictions. We extend this approach further by producing top-$N$ hypotheses from each level-0 model in the ensemble to feed as input to level 1; as we show in Figure~\ref{fig:top-8}, the set of top-$N$ predictions (when $N > 1$, compared to the set of top-1 predictions) has a much better chance of including the correct answer.

To learn level-0 models, we fine-tuned various state-of-the-art PCE models using the provided SQuAD2.0 starter files~\cite{squadStarter} to follow proper data hygiene practices given the dev-test split. To learn the meta-model in level 1, we gave it a classification task: We selected the top-$N$ hypotheses produced by $M$ level-0 models, computed the F1 score distribution for the resulting $M \times N$ hypotheses given the ground-truth answers, and then asked the meta-model to predict the F1 score distribution for a set of $M \times N$ hypotheses; the ensemble's predicted answer is the $\mathrm{argmax}$ of the predicted F1 scores.

For level 0, we selected the top-8 hypotheses produced by two models: \texttt{albert-xxlarge-v1} and \texttt{roberta-base}~\cite{pretrained}. The former is the best performing level-0 model (on the dev set) and doubles as the baseline: It achieved an EM score of 85.966 and an F1 score of 88.945 while its top-8 scores were 94.965 (EM) and 95.473 (F1). We picked \texttt{roberta-base} because it was the best performing model outside of the ALBERT family of models~\footnote{XLNet's results (see Table~\ref{tab:single_models}) were ignored because they were too low, possibly due to a bug in the transformers library's implementation: \url{https://github.com/huggingface/transformers/issues/2651}} and because of its relatively smaller size (125M parameters, compared to 223M parameters in \texttt{albert-xxlarge-v1}); notably, its top-1 scores were relatively low: 75.337 (EM) and 78.683 (F1); however, its top-8 scores were 94.373 (EM) and 95.372 (F1). More importantly, combining the two set of hypotheses yielded the synergistic scores of 98.289 (EM) and 98.539 (F1), thanks to the complementary nature of the blend, which we can use as an oracle for the meta-model in level 1. See Table~\ref{tab:single_models} for a detailed comparison of PCE models' scores.

\begin{figure}[ht]
    \centering
    \includegraphics[width=1\textwidth]{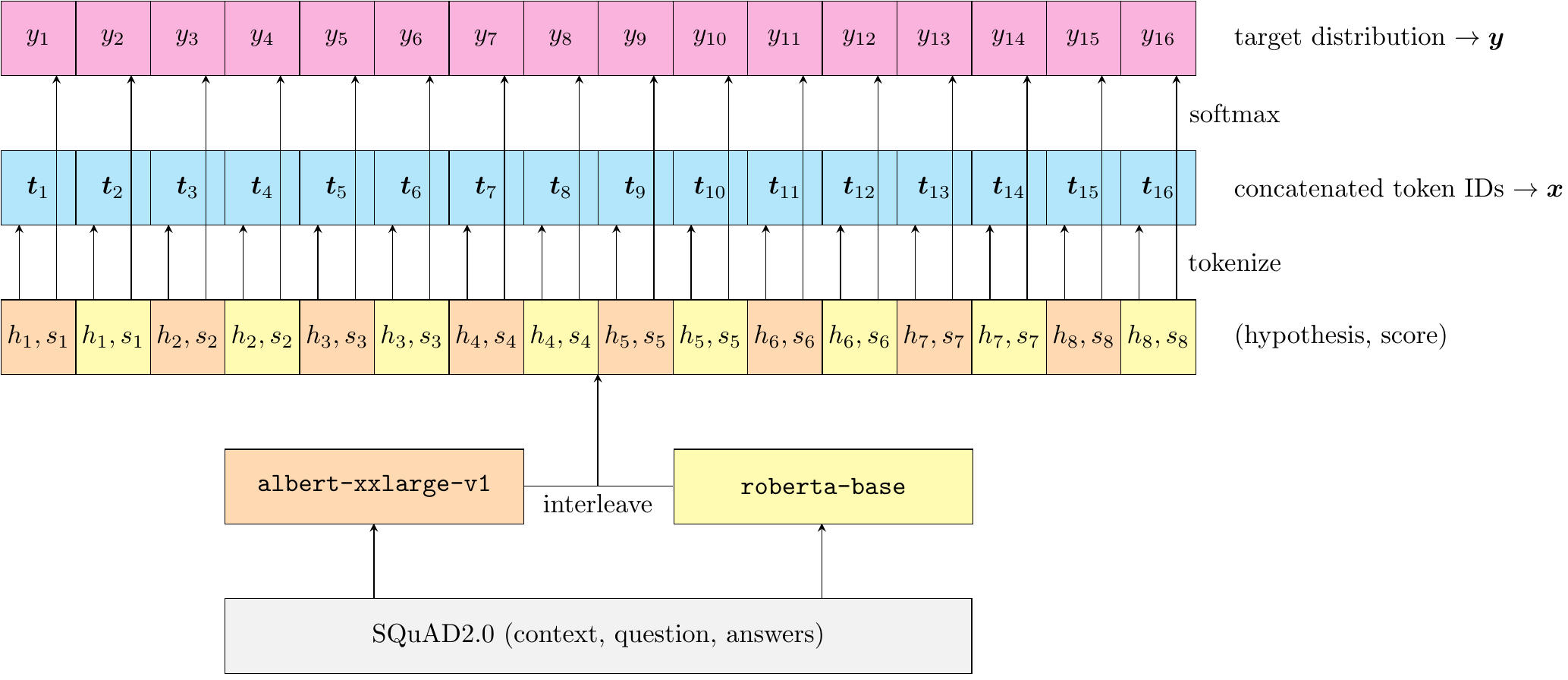}
    \caption{A simplified flow of input $\bm{x}$ and output $\bm{y}$ data processing to learn a meta-model in level 1.}
    \label{fig:meta-model-input-output}
\end{figure}

For level 1, 16 hypotheses $\bm{h}$ (top-8 from two models) and their respective F1 scores $s_i \in [0, 1]$ (with respect to the ground truth) were interleaved to make up a vector which preserved their original relative order. When a model produced fewer than 8 hypotheses, they were padded with a special token (\texttt{<ap>}) and a respective F1 score of 0; the percentages of padded examples in the train, dev, and test datasets were 10.9\%, 0\%, and 11.4\%, respectively. The no-answer hypothesis also gets its own special token (\texttt{<na>}). We prepended a unique prefix token to each of the 16 hypotheses to identify their positions (\texttt{<h1>, <h2>, ..., <h16>}), then encoded each input string using a T5 tokenizer~\cite{2019t5, Wolf2019HuggingFacesTS}, which was reconfigured to learn about the newly added special tokens, and obtained 16 vectors of token identifiers $\bm{t}$. Each $\bm{t}$ vector is of size 16; tokens were truncated if they exceeded the vector's size (the maximum length for an answer was set to 30 in level 0); conversely, tokens were padded with a padding token (\texttt{<pad>}) if they were fewer than 16. We concatenated the 16 vectors to represent the input hypotheses to the meta-model as an input vector $\bm{x}$. The target $\bm{y}$ was designed to be the distribution of scores $\bm{s}$ after applying the $\mathrm{softmax}$ function. See Figure~\ref{fig:meta-model-input-output} for an illustration.

Our best-performing meta-model used \texttt{t5-small}~\cite{pretrained, 2019t5} to generate contextual embeddings, each is of size 512, for tokens in $\bm{x}$. The PCE model in level 1 (in this case, \texttt{t5-small}), was not fine-tuned for any specific task before using it as a meta-model module; we allowed the PCE model's parameters to change given the new task and special tokens we introduced in level 1; we also resized the PCE model to include said tokens in its vocabulary. T5 was our default choice for a PCE model in level 1 given the flexibility it provides for arbitrary special tokens, state-of-the-art performance at various NLP tasks, and being different from the two PCE models we fine-tuned in level 0: \texttt{albert-xxlarge-v1} and \texttt{roberta-base}; empirical results, discussed in Section ~\ref{sec:experiments}, supported our default choice.

\begin{figure}[ht]
    \centering
    \includegraphics[width=1\textwidth]{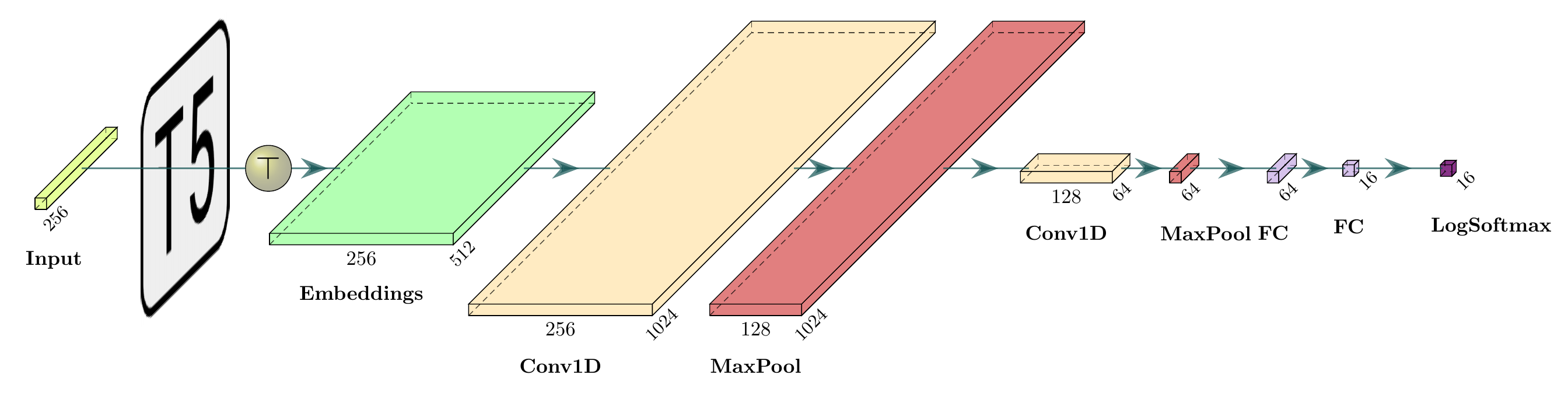}
    \caption{Architecture of the best-performing meta-model in level 1.}
    \label{fig:meta-model-arch}
\end{figure}

The best-performing meta-model architecture mapped the embeddings matrix to the target distribution $\bm{y}$ using a series of transformations: transposing dimensions so that the embedding size is the number of input channels for the first convolution; two convolutional blocks, each applied a 1D convolution then max-pooling, with 1024 and 64 output channels, respectively; followed by two fully connected layers, with 64 and 16 nodes, respectively; and finally a log-softmax output layer, which predicts the log-probabilities $\bm{\hat{y}}$ for the 16 input hypotheses. We picked the Kullback–Leibler (KL) divergence loss~\cite{kullback1951information} with a summative reduction as the cost function to minimize: $$D_{\mathrm{KL}}(\bm{\hat{y}} \parallel \bm{y}) = \sum_{i} \hat{y}_{i} \log \frac{\hat{y}_{i}}{y_{i}}$$

The KL divergence loss impels the meta-model to learn to predict log-probability scores for a mix of --- potentially multiple --- correct, partially correct, and incorrect hypotheses in the input $\bm{x}$. We used the rectified linear unit (ReLU) activation for nonlinearity~\cite{glorot2011deep}. For optimization, we picked Adam with a dynamic learning rate that gets reduced when the dev F1 score plateaus~\cite{kingma2014adam}. We implemented the ensemble using PyTorch, parts of the provided starter code, and PCE models from the transformers library~\cite{pytorch, squadStarter, Wolf2019HuggingFacesTS}.

\section{Experiments}
\label{sec:experiments}
\subsection{Evaluation Methods}
To evaluate the ensemble, we used the same evaluation metrics as the SQuAD2.0 task: Exact Match (EM) and F1 scores.
\begin{itemize}
    \item \textbf{EM:} is a binary measure of correctness --- the score is 1 if the predicted answer exactly matches the ground truth; otherwise, it's 0.
    
    \item \textbf{F1:} is the harmonic mean of precision and recall of words in the predicted answer given the ground truth.
    $F1 = 2 \times \mathrm{precision} \times \mathrm{recall} \mathbin{/} (\mathrm{precision} + \mathrm{recall})$
\end{itemize}

Predictions and ground-truth answers are normalized before evaluation: We convert them to lowercase and remove punctuation, articles, and extraneous whitespaces. For each question in the evaluation dataset, multiple ground-truth answers are supplied; we pick the best score for a prediction given the ground-truth answers to the respective question, then average the scores across the entire dataset. We use the performance, EM and F1 scores, of the best model in the ensemble as a baseline to outperform.

\begin{figure}[ht]
    \centering
    \includegraphics[width=1\textwidth]{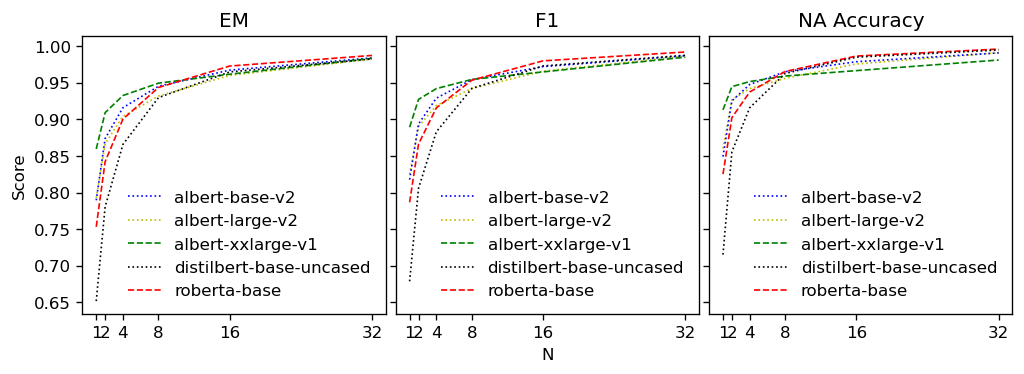}
    \caption{Top-$N$ EM, F1, and No-Answer Accuracy scores of level-0 models. Using the top-8 answers as input to the meta-model provides a good lift in metrics (at a relatively small $N$).}
    \label{fig:top-8}
\end{figure}

To evaluate the performance of level-0 models and their respective top-$N$ predictions, we used the same metrics listed above; we evaluated the models at $N \in \{1, 2, 4, 8, 16, 32\}$ and picked the best score out of the $N$ results for each metric. We used the results of level 0 to curate the input to level 1 (the choice of models and the value of $N$); see Figure~\ref{fig:top-8} for a comparison of the results at a glance.

\subsection{Data}
For level 0, we used the given SQuAD2.0 starter files~\cite{squadStarter} (with half of the official dev dataset repurposed as the test dataset) to fine-tune PCE models. The train/dev/test splits consist of 129,941/6,078/5,915 examples, respectively. Each input consists of a question and a context paragraph pair; each output represents the span of text in the context paragraph that answers the respective question or indicates the lack of such answer. Almost a third of the questions in the dataset are unanswerable given their respective contexts~\cite{rajpurkar2018know}.

For level 1, we selected two models (\texttt{albert-xxlarge} and \texttt{roberta-base}) from level 0 to contribute to the ensemble; we used the top-8 predictions and their respective F1 scores for the train, dev, and test datasets from each model in level 0 to create the respective train, dev, and test datasets required to learn the neural meta-model in level 1. We ignored the targets (answers in level 0 and F1 score distributions in level 1) for the test data sets.

\subsection{Experimental Details}
\subsubsection{Fine-Tuning Level-0 PCE Models}
Table~\ref{tab:single_models} shows detailed configurations and results of fine-tuning level-0 PCE models~\footnote{Fine-tuned models are available at  \url{https://huggingface.co/models?search=elgeish}} using SQuAD2.0. XLNet models were not considered for the ensemble because they scored too low, possibly due to a bug in the transformers library’s implementation of XLNet~\cite{bugXLNet}. We used \texttt{albert-xxlarge-v1} since it outperforms v2, thanks to carefully tuned hyperparameters~\cite{githubALBERT}. The performance of our \texttt{albert-xxlarge-v1} model (85.97 EM and 88.95 F1 on our dev set; 86.64 EM and 89.76 F1 on our test set) is slightly worse than what's reported in~\cite{Lan2020ALBERT} and the official SQuAD2.0 leaderboard (87.4 EM and 90.2 F1 on the official dev set; 88.107 EM and 90.902 F1 on the official test set). The discrepancies can be explained by the differences in the evaluation datasets (only using half of the official dev set for model selection) and under-training (due to the limited training budget).

\begin{table}[ht]
    \centering
    \setlength{\tabcolsep}{3pt}
    \begin{tabular}{l l l l l l l l}
        &
        \rot{\textbf{xlnet-base-cased}} &
        \rot{\textbf{xlnet-large-cased}} &
        \rot{\textbf{distilbert-base-uncased}} &
        \rot{\textbf{roberta-base}} &
        \rot{\textbf{albert-base-v2}} &
        \rot{\textbf{albert-large-v2}} &
        \rot{\textbf{albert-xxlarge-v1}} \\
        \toprule
        \textbf{Gradient Accumulation Steps} & 1 & 24 & 24 & 24 & 24 & 1 & 24 \\
        \textbf{Learning Rate} & 3e-5 & 3e-5 & 3e-5 & 3e-5 & 3e-5 & 3e-5 & 3e-5 \\
        \textbf{Max Answer Length} & 30 & 30 & 30 & 30 & 30 & 30 & 30 \\
        \textbf{Max Query Length} & 64 & 64 & 64 & 64 & 64 & 64 & 64 \\
        \textbf{Max Sequence Length} & 384 & 384 & 384 & 384 & 384 & 384 & 512 \\
        \textbf{Training Epochs} & 4 & 3 & 4 & 4 & 3 & 5 & 4 \\
        \textbf{Training Batch Size} & 16 & 4 & 32 & 16 & 8 & 8 & 1 \\
        \midrule
        \textbf{Top-1 Dev EM Score} & 35.604 & 38.549 & 65.186 & 75.337 & 78.957 & 79.286 & 85.966 \\
        \textbf{Top-1 Dev F1 Score} & 40.339 & 42.152 & 67.890 & 78.683 & 81.789 & 82.525 & 88.945 \\
        \textbf{Top-8 Dev EM Score} & --- & --- & 92.991 & 94.373 & 94.538 & 93.205 & 94.965 \\
        \textbf{Top-8 Dev F1 Score} & --- & --- & 94.245 & 95.372 & 95.447 & 94.230 & 95.473 \\
        \bottomrule \\
    \end{tabular}
    \caption{Fine-tuning experiments and notable results for level-0 PCE models.}
    \label{tab:single_models}
\end{table}

\subsection{Level-1 Meta-Model}
We explored various data representations, embedding models, meta-model architectures, and other hyperparameters in order to improve performance:
\paragraph{Representation design choices.} \begin{enumerate*}[label=(\roman*)]
  \item assigning each hypothesis in the input its own unique special token as a prefix vs. using a single separator token
  \item tokenizing and padding each hypothesis and token pair separately vs. concatenating the pairs then tokenizing with padding at the end
  \item PCE models for language representation: \texttt{t5-small}, \texttt{t5-base}, \texttt{xlnet-large-cased}, and \texttt{albert-xlarge-v2}
  \item adding the question to the input with its own unique prefix token
  \item maximum answer length values of 16, 24, and 32; the latter doesn't truncate any input since the maximum length expected is 31: 30 from the output of level-0 models and the prefix token
  \item penalizing the model further when it erroneously picks an answer for an unanswerable question or a no-answer hypothesis for an answerable one; incorrect hypotheses, with respect to the answerability condition of the given question, are assigned scores of -1 instead of their respective F1 scores; we refer to this variant below in Table~\ref{tab:test_leaderboard} as biased $\bm{y}$; see Figure~\ref{fig:target-distribution} for an example.
\end{enumerate*}

\paragraph{Architecture design choices.} To transform the embeddings, we started with a single fully connected hidden layer then increased the complexity of the model by adding more nodes and layers (up to 4) as needed. We also experimented with 1D convolutions (up to 3 blocks) followed by fully connected layers (up to 2) in the same manner. In addition, we explored the addition of a self-attention layer (that attends to the embeddings) to the meta-model variants above.

\paragraph{Regularization techniques.} \begin{enumerate*}[label=(\roman*)]
  \item dropout for hidden layers that transform the embeddings, with hand-tuned values for the dropout probability $p \in \{0.0, 0.1, 0.2, 0.3, 0.5, 0.6, 0.8\}$
  \item dropout for the embeddings with $p \in \{0.0, 0.2\}$
  \item adding Gaussian noise, with $\sigma = 0.1$, to the embeddings
\end{enumerate*}

\paragraph{Optimization design choices.} We tested multiple values for Adam's initial learning rate: 0.01, 0.001, 0.0005, and 0.0001. For the reduce-on-plateau scheduler, we experiments with 10 and 3 as values for the patience parameter.

Given the large number of combinations, we tested variables with the highest expected impact first, one at a time. Then, we used the newly obtained best-performing configuration as a baseline for the next round of experiments. To mitigate myopic design choices, certain variants (e.g., convolutional and non-convolutional architectures) were retested with addition or ablation of other significant variables. It's worth noting that EM and F1 dev scores moved in the same direction (increasing or decreasing together) but not necessarily with the same ratio; when a trade-off was needed for model selection, we picked variants that better improved the F1 score.

\subsection{Results}
On the PCE leaderboard, we reported the following test scores:
\begin{table}[ht]
    \centering
    \begin{tabular}{lcc}
    \textbf{Model} & \textbf{Test EM Score} & \textbf{Test F1 Score} \\
    \toprule
    Baseline (\texttt{albert-xxlarge-v1}) & 86.644 & 89.760 \\
    \texttt{t5-small}, 0.2 dropout, 2 FC layers (256 $\rightarrow$ 16), biased $\bm{y}$ & 87.050 & 90.239 \\
    Best-performing ensemble (described in Section~\ref{sec:approaches}) & \textbf{87.117} & \textbf{90.306} \\
    \bottomrule \\
    \end{tabular}
    \caption{Test scores we reported on the PCE leaderboard.}
    \label{tab:test_leaderboard}
\end{table}

Given the use of \texttt{albert-xxlarge-v1} and scores of the level-1 oracle (the combined output of level-0 hypotheses) of 98.289 (EM) and 98.539 (F1), we expected gains comparable to the ones obtained by ALBERT's ensemble (1.3 F1 points), even though it averages the scores of 6 to 17 models. Error analysis of output samples shows that the predominant error class is answering an unanswerable question; this could be explained by the design choices of the target distribution $\bm{y}$ and the KL divergence loss function. A potential improvement here might be obtained by breaking down the level-1 meta-model into a pipeline of two tasks: binary classification (answerable vs. unanswerable) and multiclass classification (which answer to output) when the first stage indicated that the question is answerable. Alternatively, the two tasks can be combined in a multi-objective optimization setting, which can make use of the no-answer scores (null odds) generated by the transformers library.
\section{Analysis}
Using the dev set, we found 162 disagreements (out of 6,078 examples) between the normalized answers of the ensemble and the baseline model; 73\% of the disagreements were attributed to a difference in opinion regarding the answerability of questions. In addition, we manually inspected 50 randomly chosen results from the test set (without answers) to better understand the ensemble's output, compared to the baseline. We also repeated the same analyses for two variants of the ensemble to include a qualitative factor into the model selection process. Given our own on-the-fly answers to questions in the random test sample, we found that the best-performing ensemble, unsurprisingly, made the fewest mistakes and when it disagreed with the baseline model, other ensemble variants agreed.

Below, we list a few examples of disagreements between the best-performing ensemble and the baseline model.

\begin{figure}[ht]
    \centering
    \begin{tabular}{p{0.9\linewidth}}
        \toprule
        \textbf{Paragraph:} "The further decline of Byzantine state-of-affairs paved the road to a third attack in 1185, when a large Norman army invaded Dyrrachium, owing to the betrayal of high Byzantine officials. Some time later, Dyrrachium—one of the most important naval bases of the Adriatic—fell again to Byzantine hands." \\ \\
        \textbf{Question 1:} "Who betrayed the Normans?" \\
        \textbf{Ground Truth Answer:} <No Answer> \\
        \textbf{Baseline's Answer:} high Byzantine officials \\
        \textbf{Ensemble's Answer:} <No Answer> \\
        \bottomrule
    \end{tabular}
    \vspace{1em}
    \caption{An example of the baseline model incorrectly answering an unanswerable question while the best-performing ensemble picked the correct no-answer hypothesis, which was ranked 2nd by the baseline model (\texttt{albert-xxlarge}) and 3rd by \texttt{roberta-base} in level 0.}
\end{figure}

\begin{figure}[ht]
    \centering
    \begin{tabular}{p{0.9\linewidth}}
        \toprule
        \textbf{Paragraph:} "If the input size is n, the time taken can be expressed as a function of n. Since the time taken on different inputs of the same size can be different, the worst-case time complexity T(n) is defined to be the maximum time taken over all inputs of size n..." \\ \\
        \textbf{Question 1:} "How is worst-case time complexity written as an expression?" \\
        \textbf{Ground Truth Answer:} T(n) \\
        \textbf{Baseline's Answer:} T(n) \\
        \textbf{Ensemble's Answer:} <No Answer> \\
        \bottomrule
    \end{tabular}
    \vspace{1em}
    \caption{An example of the best-performing ensemble incorrectly picking a no-answer hypothesis, which was ranked as the top answer by \texttt{roberta-base} in level 0. It's worth noting that 7 hypotheses, out of the 16 in level 0, had an F1 score of 0; that's expected given the very short ground-truth answer and hypotheses that don't overlap with the ground truth.}
\end{figure}

\begin{figure}[ht]
    \centering
    \begin{tabular}{p{0.9\linewidth}}
        \toprule
        \textbf{Paragraph:} "The most commonly used reduction is a polynomial-time reduction. This means that the reduction process takes polynomial time. For example, the problem of squaring an integer can be reduced to the problem of multiplying two integers. This means an algorithm for multiplying two integers can be used to square an integer. Indeed, this can be done by giving the same input to both inputs of the multiplication algorithm. Thus we see that squaring is not more difficult than multiplication, since squaring can be reduced to multiplication." \\ \\
        \textbf{Question 1:} "According to polynomial time reduction squaring can ultimately be logically reduced to what?" \\
        \textbf{Ground Truth Answer:} multiplication \\
        \textbf{Baseline's Answer:} multiplication \\
        \textbf{Ensemble's Answer:} multiplication, since squaring can be reduced to multiplication. \\
        \bottomrule
    \end{tabular}
    \vspace{1em}
    \caption{An example of the best-performing ensemble picking a reasonable answer; yet, it's still penalized given the ground-truth answers, which agreed on the single-word answer --- an expected case given the preference of short answers over long ones in SQuAD2.0. The ensemble ignored the two top level-0 answers, which are correct.}
\end{figure}
\section{Conclusion}
Ensemble learning is a tremendously useful technique to improve upon state-of-the-art models; it helps models generalize better and overcome their weaknesses. In a stacking-ensemble setting, heterogeneous level-0 models can complement each other like a gestalt --- when blended properly, the ensemble outperforms the best model in level 0. Our best-performing ensemble combined the top-8 hypotheses from each of \texttt{albert-xxlarge-v1} and \texttt{roberta-base} in level 0; incorporated \texttt{t5-small} to generate contextual embeddings; transformed the embeddings through a CNN-based meta-model; and achieved relative improvements of 0.55\% for EM and 0.61\% for F1 scores, compared to the baseline performance of the best model in level 0. We believe there's still room for improvement and future work that explores experiments such as: \begin{enumerate*}[label=(\roman*)]
  \item testing various combinations of level-0 models, including non-PCE ones, and values of $N > 8$ that provide better input to level 1
  \item backpropagating errors to level-0 models while training the meta-model in level 1
  \item data augmentation using distractors (e.g., incorrect answers) for better generalization
  \item multi-objective optimization that takes into account a linear combination of EM and F1 metrics
  \item thorough hyperparameter optimization, etc.
\end{enumerate*} Finally, we conclude that stacking ensembles can improve upon state-of-the-art SQuAD2.0 systems when properly blended with a neural meta-model. Moreover, it can benefit single models and other ensembles alike, since it's agnostic to the source of the meta-model's input.

\newpage
\bibliographystyle{unsrt}
\bibliography{references}
\newpage
\appendix
\section{Appendix - Supplemental Figures and Tables}
\label{app:a}

\begin{table}[ht]
    \centering
    \setlength{\tabcolsep}{3pt}
    \begin{tabular}{lcccccccccccc} \toprule \textbf{Dev Metric} & \multicolumn{6}{c}{\textbf{EM}} & \multicolumn{6}{c}{\textbf{F1}} \\
    \textbf{Model \textbackslash \ $N$} & 1 & 2 & 4 & 8 & 16 & 32 & 1 & 2 & 4 & 8 & 16 & 32 \\
    \midrule
    \textbf{albert-large-v2 } & 79.3 & 86.7 & 90.4 & 93.2 & 96.0 & 98.3 & 82.5 & 89.0 & 91.9 & 94.2 & 96.6 & 98.7 \\
    \textbf{albert-base-v2 } & 79.0 & 87.4 & 91.6 & 94.5 & 96.7 & 98.4 & 81.8 & 89.4 & 92.9 & 95.4 & 97.2 & 98.7 \\
    \textbf{albert-xxlarge-v1 } & 86.0 & 90.9 & 93.3 & 95.0 & 96.2 & 98.4 & 88.9 & 92.7 & 94.2 & 95.5 & 96.5 & 98.5 \\
    \textbf{roberta-base } & 75.3 & 84.2 & 90.0 & 94.4 & 97.3 & 98.7 & 78.7 & 86.6 & 91.6 & 95.4 & 98.0 & 99.2 \\
    \textbf{distilbert-base-uncased} & 65.2 & 78.1 & 86.6 & 93.0 & 96.5 & 98.3 & 67.9 & 80.6 & 88.3 & 94.2 & 97.3 & 98.8 \\
    \bottomrule \\
    \end{tabular}
    \caption{Detailed top-$N$ results for notable level-0 PCE models.}
\end{table}

\begin{figure}[ht]
    \begin{subfigure}{0.495\textwidth}
    \includegraphics[width=0.9\linewidth]{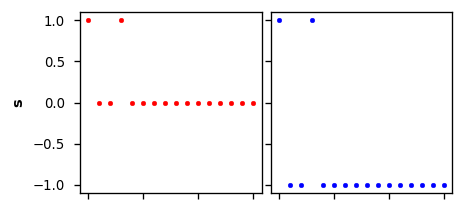}
    \caption{$\bm{s}$ as F1 scores (red) and adjusted scores (blue)}
    \end{subfigure}
    \begin{subfigure}{0.495\textwidth}
    \includegraphics[width=0.9\linewidth]{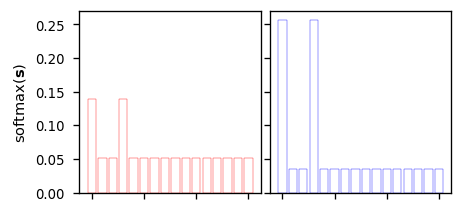}
    \caption{The two variants of $\bm{s}$ after applying $\mathrm{softmax}$}
    \end{subfigure}
\caption{Example of target variants when a question is unanswerable; only the first and the fourth hypotheses are correct (no answer). The blue variant achieved a slightly better dev-set performance.}
\label{fig:target-distribution}
\end{figure}

\end{document}